 \documentclass[letterpaper, 10 pt, conference]{ieeeconf}  

\IEEEoverridecommandlockouts                              

\usepackage{cite}
\usepackage{amsmath,amssymb,amsfonts}
\usepackage{algorithmic}
\usepackage{graphicx}
\usepackage{textcomp}
\usepackage{xcolor}
\usepackage{multirow}
\usepackage{bm}
\usepackage{booktabs}
\usepackage{float}
\usepackage{tabularx}

\title{\LARGE \bf
Base Placement Optimization for Coverage Mobile Manipulation Tasks
}

\author{Huiwen Zhang$^{1,2}$ Kai Mi$^{1}$ and Zhijun Zhang$^{2}$
\thanks{$^{1}$Huiwen Zhang is with the Institute of CVTE Research and the School of Automation Science and Engineering, South China University of Technology, Guangdong, China. Kai Mi is with the Institute of CVTE Research
        {\tt\small huiwen@mail.ustc.edu.cn}}%
\thanks{$^{2}$Zhijun Zhang is with the School of Automation Science and Engineering, South China University of Technology, Guangzhou, China
        {\tt\small auzjzhang@scut.edu.cn}}%
}

\begin{document}

\maketitle
\thispagestyle{empty}
\pagestyle{empty}

\begin{abstract}
 Base placement optimization (BPO) is a fundamental capability for mobile manipulation and has been researched for decades. However, it is still very challenging for some reasons. First, compared with humans, current robots are extremely inflexible, and therefore have higher requirements on the accuracy of base placements (BPs). Second, the BP and task constraints are coupled with each other. The optimal BP depends on the task constraints, and in BP will affect task constraints in turn. More tricky is that some task constraints are flexible and non-deterministic. 
Third, except for fulfilling tasks, some other performance metrics such as optimal energy consumption and minimal execution time need to be considered, which makes the BPO problem even more complicated.
In this paper, a Scale-like disc (SLD)  representation of the workspace is used to decouple task constraints and BPs. To evaluate reachability and return optimal working pose over SLDs, a reachability map (RM) is constructed offline. In order to optimize the objectives of coverage, manipulability, and time cost simultaneously, this paper formulates the BPO  as a multi-objective optimization problem (MOOP). Among them, the time optimal objective is modeled as a traveling salesman problem (TSP), which is more in line with the actual situation. 
The evolutionary method is used to solve the MOOP. Besides,  to ensure the validity and optimality of the solution, collision detection is performed on the candidate BPs, and solutions from BPO are further fine-tuned according to the specific given task.  Finally, the proposed method is used to solve a real-world toilet coverage cleaning task. Experiments show that the optimized BPs can significantly improve the coverage  and efficiency of the task.
\end{abstract}

\section{INTRODUCTION}
Planning a reasonable BP  has a very important impact on the  sequential task execution. For humans, this task can be accomplished very easily for two reasons. First, humans have very dexterous motor skills, so they have  greater tolerance for base placement errors. Second, human beings have accumulated a lot of daily operation experience, which can make them easily predict a  reasonable BP according to the task constraints. However, it is a non-trivial task for robots. This is because current robots are extremely inflexible compared with human beings. Although mobile manipulators have more degrees of freedom (DoFs), they are generally independently controlled. Therefore, the robot is very sensitive to BPs. On the other hand, the BP and task constraints are coupled together, and the optimal BPs depend on the task constraints, while task constraints are usually decided by the current BP. 
In addition, the robot is often expected to meet some other performance standard such as optimal energy and minimal running time, which makes the BPO problem even more complicated.

\begin{figure}
    \centering
    \includegraphics[width=\linewidth]{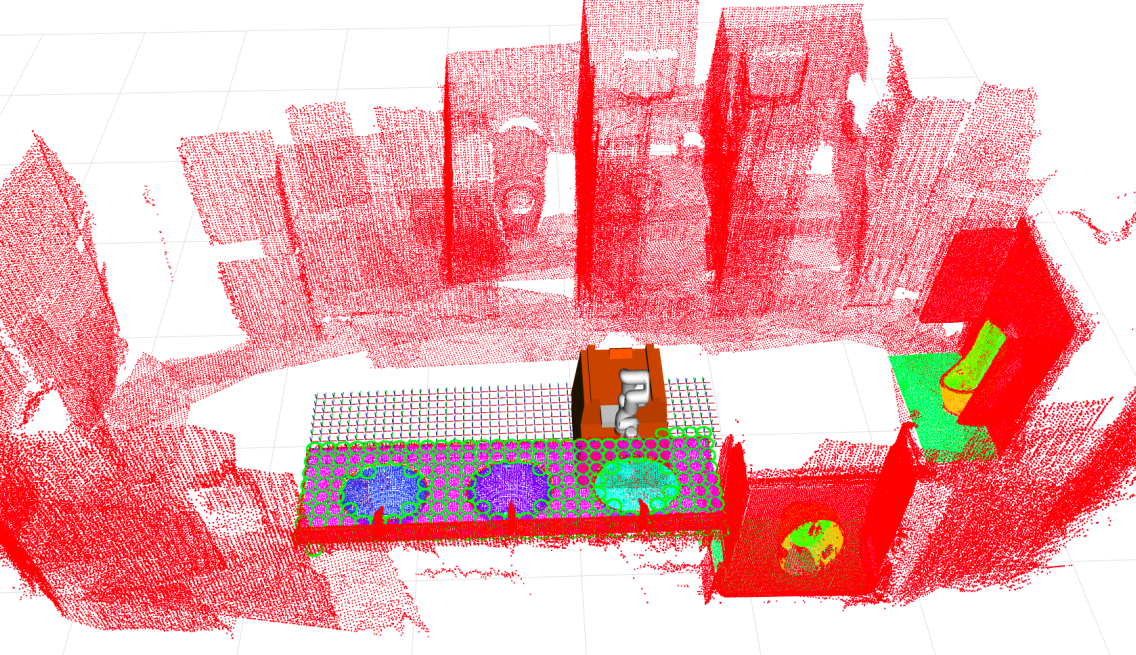}
    \caption{BPO for the coverage cleaning task. The picture shows a reconstructed point cloud model of the toilet. To fulfill the washbasin and urinal cleaning task in a large range, the robot needs to stop at several optimized placements and ensures the overall coverage is maximal. Circles with green color represent the SLD model used to approximate the surface of washbasin. Axes arrays on the ground stand for the sampled candidate BPs.}
    \label{fig:task}
\end{figure}

Coverage planning is a very important task for mobile manipulation\cite{galceran_survey_2013}, and a typical application is the cleaning task, such as table cleaning, garbage disposal\cite{kabir_Automated_2017, yin_Table_2020, martinez_Planning_2015} and steel bridge maintenance\cite{paul_Robotic_}.  However, most works focus on coverage path planning, ignoring the importance of the BPO problem. In addition, most of the tasks presented in the literature only involve local cleaning job and do not require full coverage. The task setting of this paper involves full coverage cleaning of a public toilet on multiple work sites, as shown in Fig.\ref{fig:task}. Therefore, the optimization of BPs become very critical for task performance.

Usually, BPO can be solved by inverse kinematics. The mainstream idea is to  establish a mapping  between the joint space and the task space offline, such as reachability map \cite{porges_Reachability_2014} and capability map\cite{zacharias_Capturing_}, and then obtain the optimal BP by querying the reachability inversion\cite{vahrenkamp_Robot_2013a}. The difference lies in the representation structure  of the map. 
However, the majority of works are focused on grasping or planning tasks, so only the local reachability  is concerned; Secondly, the task setting  only cares about the reachability and manipulability  on a single placement, and does not consider global optimal solutions for sequential tasks.  Besides, the trajectory of the cleaning task does not have strict pose constraints, so there may be an infinite number of valid reachable configurations. In this case, solving the inverse kinematics is very time-consuming.

In this paper, a SLD-based representation of the workspace is used for decoupling of task constraints and BPO. It discretizes the workspace into several connected nodes, and evaluates the position of the chassis by calculating the accessibility and manipulability of the robot over all nodes. The RM has nothing to do with the specific task settings, so it can be well decoupled from the task constraints. In order to get inverse reachability and optimal working pose, a hash map which stores  manipulability, collision information and  corresponding configurations are used.
In order to consider the objectives of coverage, manipulability, and time cost simultaneously, this paper formulates the BPO problem as a MOOP, and  uses the evolutionary algorithm to find optimal solutions. The idea of this paper is similar to \cite{hassan_Enabling_2018}, the difference is that \cite{hassan_Enabling_2018} solves the BPO problem for multi-robot collaborative coverage planning, and this paper aims at  single robot's coverage planning on multiple placements. Secondly, collision detection is introduced in computing favoured base placements (FBP). Experiments show that this process will directly affect the feasibility of the optimal BP. Besides, we formulate the time optimal objective as a TSP, which is more reasonable.

The contributions of this paper are as follows:
\begin{itemize}
    \item Representing the workspace with SLD, which can  decouple task constraints and BPs, and a new RM is used to obtain the optimal working pose for tasks with soft constraints.
    \item Modelling the sequential coverage cleaning task as a MOOP. The  coverage, manipulability and time cost can be easily considered in a single framework. 
    \item The proposed method is validated on a coverage cleaning task. The experiments show that the proposed method can greatly increase the coverage and robustness of the mobile manipulation task.
\end{itemize}

\section{Related work}
The approaches for solving optimal BPs can be divided into three categories, namely,  RM based, neural networks (NN) based, and  optimization based. The offline RM generally uses a certain kind of data structure and is constructed through inverse kinematics. Finally, the optimal BPs are obtained through retrieval or heuristic search. \cite{makhal_Reuleaux_2017} presents a classic way to construct a RM. First, the task space is discretized into voxels,  and a sphere is established for each voxel, then some poses are sampled on the surface of the sphere. The inverse kinematics is performed on each sphere pose to evaluate reachability. If it is reachable,  store the data to the RM. The inverse reachability map (IRM) is used when solving for the BP. \cite{vahrenkamp_Robot_2013a} considers quality information such as manipulability and self-distance on the basis of RM, and can return the distribution of the base position. Paper \cite{zacharias_Capturing_} uses directional structures to represent reachability information. To speed up offline map construction, \cite{porges_Reachability_2014} proposes a hybrid approach. In order to consider both the navigation cost and the manipulation cost, \cite{reister_Combining_2022} uses an online cost map and a heuristic search method for base placement determination.

The second type of approaches is learning-based. \cite{honerkamp_Learning_2021} uses reinforcement learning to learn a base movement policy that keeps the target  reachable based on the target position and the current robot's velocity. By using the reachability signal as a reward function, the policy is updated by proximal policy optimization (PPO). This method can be run online, but optimality is not guaranteed. For the grasping task in dynamic environments, \cite{akinola_Dynamic_2021} uses a signed distance field to indicate the reachability and trains a neural network to characterize the grasping quality, so that BPs with poor grasping quality can be filtered out very quickly.

The third type of methods is optimization-based, which formulates BPO as a MOOP\cite{hassan_Collaboration_2018}, and solve the BP by combining RM and evolutionary algorithms.

\section{Methodology}
\subsection{Problem description}
Suppose a robot performs a series of chores at home, such as cleaning the dining table, sink, and kitchen. The robot traverses each target in turn, stops within the appropriate range of the target, stretches out the robotic arm, and performs the covering cleaning task until all tasks are fulfilled according to the specified standards. Let $z_i=(x, y, \theta)$ represent the coordinate of a BP, and $\mathcal{T} = \{\mathcal{T}_1, \mathcal{T}_2, \cdots, \mathcal{T}_n \}$ represent the given task set. For a specific robot, the feasible working space on each state $z_i$ is expressed as $\Omega = W(z_i)$, the BPO problem refers to finding a set of optimal  $z_i$ to satisfy $\mathcal{T} \subseteq \bigcup_{i=1}^m W(z_i)$. Meanwhile, the following task metrics need to be considered:
\begin{itemize}
    \item Accessibility/Coverage. The robot needs to cover as much of the work area as possible;
    \item Minimal time consuming. Minimize the number of BPs and visit the optimal BPs in a proper sequence;
    \item Manipulability. The robot should try to avoid singular positions;
    \item Safety. Minimize the risk of robot collisions.
\end{itemize}

\subsection{Reachability representation}\label{sec:sld}
\begin{figure}[ht]
    \centering
    \includegraphics[width=0.7\linewidth]{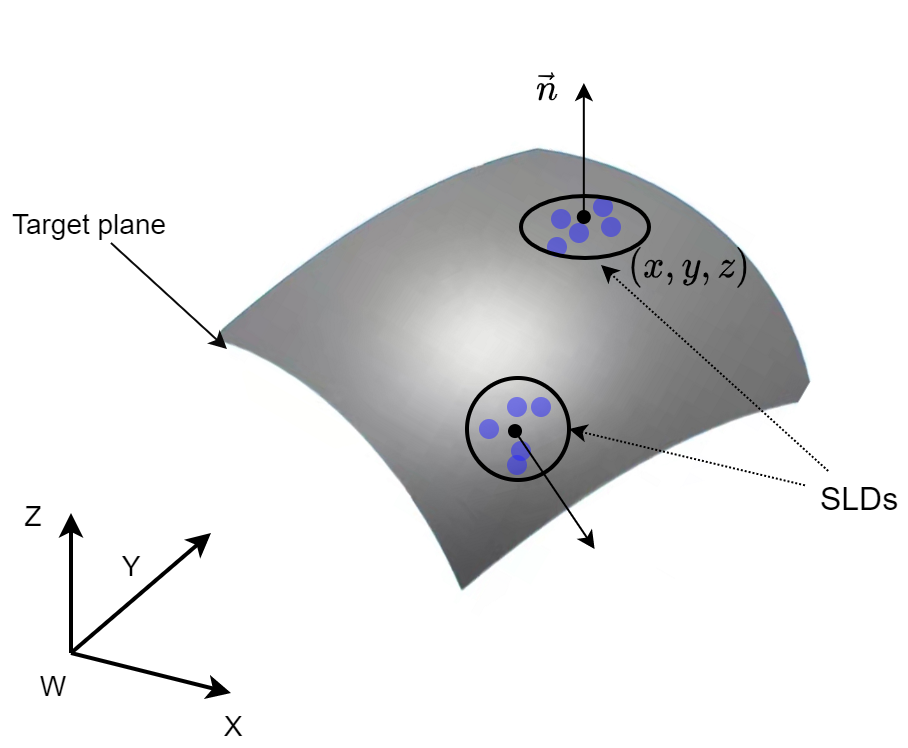}
    \caption{SLD representation of 3D surface.}
    \label{fig:sld}
\end{figure}

Usually the reachability for a given BP and task constraints are coupled together. The task constraint is generally represented by a fixed trajectory, or a pose. Reachability evaluation can be seen as solving an inverse kinematics problem if the trajectory or pose constraints are known in advance. But for some tasks, the task constraint can be a distribution over trajectories, such as the cleaning task. In this case, evaluating reachability by inverse kinematics is particularly time-consuming since a large number of candidate BPs need to be checked. The problem is more complicated if the collision constraints are also considered. Therefore, this paper evaluates the reachability based on the discrete representation of the workspace instead of the specific task at hand. Thus, no matter what the distribution of the task is, we only focus on the reachability  of each discrete area in workspace. One way to discretize an arbitrary workspace is using SLD \cite{paul_novel_2013}. The SLD method approximates the surface of the workspace by constructing  local circular surface in three-dimensional space. The obtained SLD is represented by three parameters, namely radius $r$, center coordinate $p$ and normal vector $\Vec{n}$, as shown in Fig.\ref{fig:sld}.  By computing the robot's manipulability, accessibility and other indicators over each SLD, the goodness of each BP can be verified.
\subsection{ Mathematical Modeling}
 A very important metric for cleaning tasks is coverage. In addition, it is often hoped that the robot has good dexterity in the workspace to avoid singular positions, and the robot can perform tasks with the highest efficiency and minimum time. Therefore, BPO is an optimization problem that considers multiple objectives, and the objectives are contradictory to each other. For example, in order to achieve higher coverage, the most direct way is to increase the number of BP, but the increase in the number of BP will also increase the time to fulfill the task. Therefore, this paper models the BPO problem as a MOOP, and the optimal BPs are selected from Pareto optimal front of the MOOP. The following sections will introduce the objectives considered in this paper.

\emph{Objective 1: Maximal Coverage.} Coverage indicates the range of the workspace that the robot end-effector can access. For the table cleaning task, the workspace is the entire surface of the table, and the coverage rate is the proportion of the area that can be cleaned to the total area. Based on the SLD representation introduced in Sec.\ref{sec:sld}, the coverage can be defined as:
\begin{equation}\label{eq:obj1}
    f_1(Z)=\frac{\sum_{k=1}^m N\left(z_k\right)}{C}
\end{equation}
where $Z = \{z_1, z_2,\cdots, z_m\}$ stands for the base placements, $N(z_k)$ indicates the number of SLDs that the robot can reach to at the $k$-th BP, and $C$ represents the total number of SLDs used to represent the cleaning workspace.

Given the BP $z_k$, computing the reachability to a SLD by inverse kinematics is a non-trivial task. This is mainly because the inverse calculation itself becomes very tricky if collision is considered or the configuration of the manipulator does not meet the 
Pieper criterion. What's more, for task with uncertainties, a large number of inverse operations need to be done, which is very time-consuming. To address this issue, we construct a RM $\mathcal{M}$ offline. Based on this map, the reachability of an arbitrary SLD can be retrieved very quickly, and further, the optimal reachable pose and joint configuration for the SLD can be returned simultaneously.

The  process to construct a RM $\mathcal{M}$ is summarized as follows:
1) We discretize the robot workspace into uniformly positioned cubes, expressed as $G = \text{Decompose}(l, \delta)$, where $l$ represents the maximal stretch length of the robot and $\delta$ is the resolution of discretization;
2) Discretize the joint space, then get all possible permutations $Q$ of each joint denoted, where each element in $Q$ represents a configuration of the robot;
3) For each configuration $q_j \in Q$, calculate the end effector pose $T_j$, manipulability $W_j$, and occupied cubes $G_j^o \leftarrow \text{GridOccupied}(q_j)$, where $T_j = \text{FK}(q_j)$ is calculated by the forward kinematics;
4) Get the position $p_j = (x,y,z)$ and approaching vector $\Vec{n}_j$ of the end effector from $T_j$, then the voxel index of position $p$ in $G$ is calculated by $(a,b,c) = \left(
\lfloor x / \delta \rfloor, \lfloor y / \delta \rfloor, \lfloor z / \delta \rfloor \right)$. Finally, insert key value  pair $[key = (a,b,c), value = \{p_j, q_j, n_j, W_j, G_j^o\}]$ into $\mathcal{M}$;
5) Repeat step 3 and 4 for each $q_j \in Q$.

To evaluate reachability and retrieve the optimal configuration  for  $\text{SLD}_j$, the steps are as follows:
1) First, transform the reference frame of $\text{SLD}_j$ and environmental obstacles to the robot base frame, so that task reference frame are consistent with $\mathcal{M}$;
2) Compute voxel index $(a,b,c)$ from the center coordinate of SLD ;
3) Query if there exists $key=(a,b,c)$ in $\mathcal{M}$, if not, it means the SLD is unreachable, and stop searching. Otherwise, retrieve the value corresponds to the key and  go to step 4. The value is represented as $S=\{p_k, q_k, n_k, W_k, G_k^o|k = 0,\cdots, K \}$,  $K$ is the number of valid configurations for $\text{SLD}_j$;
4) Cluster $S$ by K-means into groups $g_i$ according to the angular distance to the normal vector $n$ of $\text{SLD}_j$;
5) Return the group $g_i$ with the smallest angle distance to the approaching vector $n_j$ of the $\text{SLD}_j$;
6) Ranking $g_i$ in descending order with respect to manipulability $W_k$, and the optimal configuration is the one with maximal manipulability and collision-free. Collision checking can be verified very efficiently by using $G_k^o$.

 Given the RM $\mathcal{M}$, for each BP $z_k$ the reachability number of $N(z_k)$ is counted by following steps introduced in the last paragraph. If a SLD is reachable at different BPs, only count it once.

\emph{Objective 2: Minimal Time.} In addition to considering the coverage, it is also hoped that the used time is optimal. The work time is divided into two parts, namely the time for cleaning surface of targets and navigating between different BPs. The time for cleaning is equal to the sweep path distance  is divided by the average Cartesian velocity of the robot end-effector. To ensure full coverage, the sweep path should traverse each SLD's center. Meanwhile, to minimize sweep length, it is beneficial to visit each SLD only once. Thus, this problem can be formulated as a TSP, expressed as:
$$
   {\sum_{k=1}^m \text{TSP} \left(V\left(z_k\right) \right) } \big/ {v_{ee}}
$$
where $V(z_k)$ represents the cluster of reachable SLDs at  $z_k$, $\text{TSP}(V(z_k))$ represents the path distance solved by TSP solver,  and $v_{ee}$ represents the average Cartesian space velocity of the arm end effector. In this paper, we implement a dynamic programming method to solve the TSP.
Assuming that navigation time between two BPs is $t$, then the overall time cost is
\begin{equation}\label{eq:obj2}
    f_2(Z)= {\sum_{k=1}^m \text{TSP}(V\left(z_k\right)) } \big/ {v_{ee}}+(m-1) t
\end{equation}

\emph{Objective 3: Maximal Manipulability.}  Another indicator to consider is manipulability, which represents the dexterity of the robot in each direction of the workspace. It is denoted as
\begin{equation}\label{eq:obj3}
    f_3(Z)=\sum_{k=1}^m \sum_{j=1}^{N(z_k)} W\left(q_{k j}\right),
\end{equation}
where $W(q_{kj})$ stands for the manipulability  when the robot is in configuration $q_k$ and reaches to  $\text{SLD}_j$,  expressed as
$$
W(q_{kj})=\sqrt{\operatorname{det}\left(\mathbf{J}\left(q_{kj}\right) \cdot \mathbf{J}\left(q_{kj}\right)^\top \right)}
$$

\subsection{Multi-objective optimization}
Combining objectives \ref{eq:obj1}, \ref{eq:obj2} and \ref{eq:obj3}, the BPO problem is modeled as a multi-objective optimization problem, expressed as:
\begin{equation}\label{eq:model}
\begin{split}
    \min f(Z) &= [-f_1(Z), f_2(Z), -f_3(Z)], \\ 
    &\text{s.t.} \quad  Z \subseteq \text{FBPs}
\end{split}
\end{equation}

To solve Eq.\ref{eq:model}, there are  weighted method, $\epsilon$-constrained method, genetic algorithm, etc.  In this paper, the non-dominated sort genetic algorithm \cite{deb_fast_2002} (NSGA\uppercase\expandafter{\romannumeral2}) is used.

The chromosome in NSGA\uppercase\expandafter{\romannumeral2} represents the design variable $Z$ of the MOOP in Eq.\ref{eq:model}, and each gene on the chromosome represents a BP $z_k$. In initialization, all candidate BPs are generated with uniform sampling in preferred workspace. Each position is mapped to a integer for notation convenience and  the number 0 means no placements. For example, $Z = \{3, 0, 12\}$ means that 2 placements are required to cover the current task, corresponding to positions 3 and 12 respectively. Therefore, theoretically the algorithm will automatically remove redundant placements 
and get an optimal number of $m$. The running process of the NSGA\uppercase\expandafter{\romannumeral2} is as follows: 1) $N$ parents are randomly generated at the beginning of the algorithm; 2) Select a pair of superior children from parents by evaluating the fitness function and the crowding degree,  then perform crossover and mutation operations on children to obtain a new offspring individual; 3) Repeat step 2 until $N$ offspring individuals are obtained; 4) Merge the parent and offspring,  recalculate the fitness function and crowding degree, then select $N$ individuals from merged group as the initial population of the next generation. The process are repeated until the maximum number of generations is reached. Note that the crossover and mutation operations may lead to duplicate genes in $Z$, which means the robot will reach to a same position during the task. Obviously, it is  unreasonable, so this situation should be avoided. Besides,  adjacent placements are too close may not beneficial for exploration. This constraint can be guaranteed by limiting the sampling range when generating new individuals.
It is worth noting that the solution obtained by NSGA\uppercase\expandafter{\romannumeral2} can be guaranteed to be optimal under the SLD representation. When specific task constraints are given, a fine-turning can be  performed near the optimal solution to further improve the task performance.

\section{Experiments}

\subsection{Simulation results}
In order to verify the correctness of the proposed approach, both simulation  and physical experiments are conducted. A real toilet environment is reconstructed in the experiment. Based on the reconstructed model, the spatial position information and surface geometry information of the washbasin and urinal can be obtained. This experiment  mainly focus on the washbasin cleaning task since this task needs full coverage from different placements. First, according to the spatial position of the washbasin, candidate BPs are generated around it. In the experiment, a rectangular area in front of the washbasin  is used to generate candidate BPs. The X and Y axis of the rectangular surface are discretized with a step size of 0.05m, and the yaw angle is designed to be evenly spaced by 90 degrees, thus a total of 3213  candidate BPs are obtained. In order to speed up the optimization, the non-favoured candidate BPs are filtered  out in advance to obtain FBPs,  which should not too close or too far away from the washbasin. BPs that are too far can be filtered out by calculating the maximum reachable distance range of the robot arm. BPs that are too close are more likely to cause collisions. In this paper, both the robot and the washbasin are represented by polygons, and then collision detection for polygons in 2D plane is conducted to remove collision BPs. The size of resulting FBPs is 1840. The valid cleanning surface of the washbasin is approximated by 200 SLDs with radius set to 0.04m. The voxel resolution of the offline RM is set to 0.03m. To generate reachable pose in RM, each joint is uniformly discretized into 20 values within the limit range, After removing the self-collision configurations, a total of 172063 valid poses are obtained.
\begin{table}[]
\centering
\caption{Experiment parameters}
\label{tab:params}
\begin{tabular}{cll}
\hline
\multirow{5}{*}{NSGA\uppercase\expandafter{\romannumeral2}}               & Mutation probability    & 0.6                \\
                                    & Number of genes         & 2/3/4                 \\
                                     & Number of mutation genes         & 1                 \\
                                    & Tournament participants & 20                 \\
                                    & Population size         & 40                 \\
                                    & Iteration generations   & 80                 \\ \hline
\multirow{3}{*}{RM}                 & Voxel size              & 0.03m              \\
                                    & Pose size               & 172063             \\ 
                                    & Discretization size/DoF          & 20             \\ \hline
\multirow{4}{*}{Candidate BPs}      & Range        & 3.3m $\times$ 1.5m          \\
                                    & Resolution $(x,y,\theta)$  & (0.05m, 0.05m, 90°) \\
                                    & Number of BPs  & 3213               \\
                                    & Number of FBPs                 & 1840               \\ \hline
\multirow{2}{*}{Robot}              & Arm Cartesian velocity  & 0.1m/s            \\
                                    & Base velocity           & 0.2m/s            \\ \hline
\multirow{2}{*}{SLD}              & Radius  & 0.04m          \\
                                    & No. of SLDs           & 200        \\ \hline
\end{tabular}
\end{table}

\begin{figure}[ht]
    \centering
    \includegraphics[width=\linewidth]{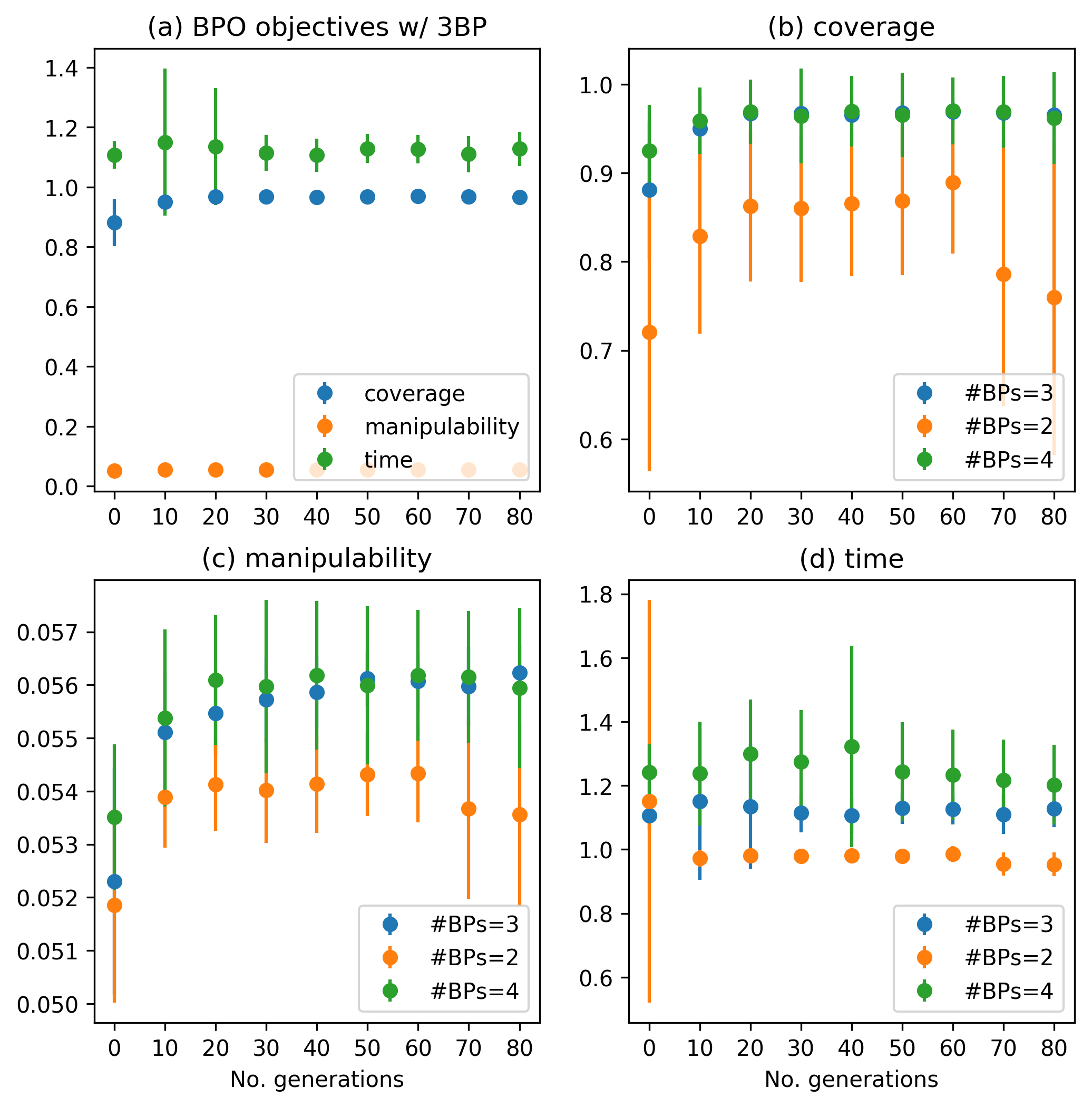}
    \caption{BPO results for toilet washbasin cleaning task.}
    \label{fig:moop}
\end{figure}

We use NSGA\uppercase\expandafter{\romannumeral2} to optimize BPs. The parameter settings of the NSGA\uppercase\expandafter{\romannumeral2} and all other parameters are summarized in Tab.\ref{tab:params}.
After running 80 generations, the obtained results are shown in Fig.\ref{fig:moop}.
Fig.\ref{fig:moop}(a) shows the mean and variance of each objective function when the initial number of BP is set to 3. As can be seen from the figure, after 10 generations, nearly 90\% coverage is achieved. As the iteration progresses, the coverage and manipulability gradually increase, and the task execution time  gradually decreases. This shows that the whole evolutionary trend is right. The final average coverage  reached more than 95\%, and the highest coverage  reached 98.5\%. Fig.\ref{fig:moop}(b), (c) and d compare the coverage, manipulability and consuming time when the initial BP number is 2, 3 and 4, respectively .Fig.\ref{fig:moop}(b) and (c) show  that  greater coverage and manipulability values are achieved if the number of BP is bigger. This is also 
in line with  common sense. When more BPs are used, the robot has better reachability, and thus the operation will be better. But at the same time, it also increases the task execution time, as shown in (d).

\begin{figure}[h]
    \centering
    \includegraphics[width=0.8\linewidth]{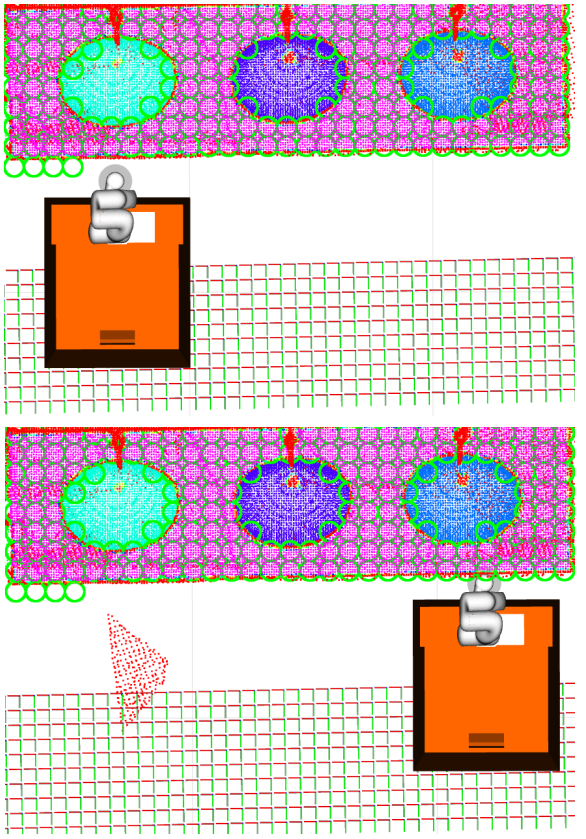}
    \caption{Optimized BPs. Axes on ground represent computed FBPs. Green circles are SLDs used for representing the washbasin. Total two BPs are optimized for washbasin task.}
    \label{fig:bps}
\end{figure}

Comparing the cases of the number of BP is equal to 3 and 4 in detail, it is found that the coverage for two cases is very close, and the manipulability is slightly better than the case of BP is 3, but the execution time is longer. This shows that when the number of BP is greater than 3, it is difficult to improve the task performance further  by using more BPs. In addition, it is worth noting that although the mean coverage and manipulability of BP is equal to 2 is not as good as that of BP is equal to 3, some solutions in Pareto front still achieve 95\% coverage and take less time. For real-world tasks, setting up more BPs in addition to increasing the navigation time, the switching of robot states between each pair of BPs will additionally increase the complexity of the task process. Therefore,  we choose the solution with the largest coverage from the Pareto optimal front obtained when BP is equal to 2 in real task execution. Specifically, 
the solution which coverage is equal to 97.5\% is selected. The corresponding coordinates of BPs are shown in the Fig.\ref{fig:bps}. It can be seen from the figure that the two BPs  are located in the center of the first and third basins, respectively, covering the area on the left and right half of the washbasin. The result is also as expected. 

\subsection{Real-world experiments}
\begin{figure}[ht]
    \centering
    \includegraphics[width=\linewidth]{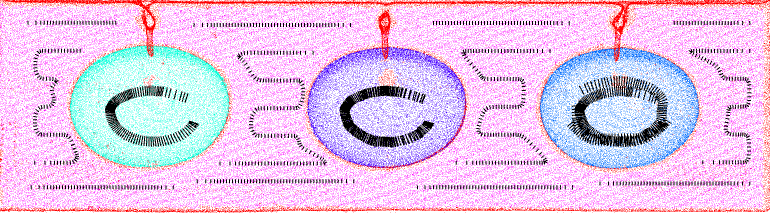}
    \caption{Example of cleaning coverage paths. Black dot lines stands for cleaning path. We don't consider path inside the basin in our experiments.}
    \label{fig:path}
\end{figure}

\begin{figure}[h]
    \centering
    \includegraphics[width=\linewidth]{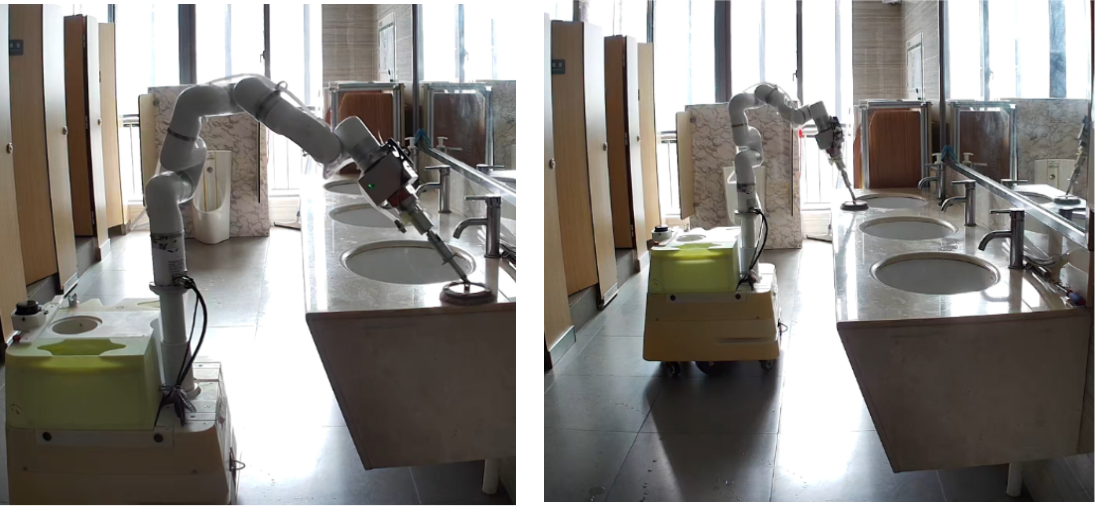}
    \caption{Snapshot of real-world washbasin cleaning task.}
    \label{fig:real_exp}
\end{figure}

Based on the  simulation results, we conduct  experiments in real-world situations. The hardware platform consists of a differential steered mobile robot and a 6-DOF collaborative robotic arm.  There is a lift  between the robotic arm and the mobile robot, as shown in Fig.\ref{fig:real_exp}. We use Nvidia Xavier as our computing platform. The arm controller, local localization  and online motion planning algorithm run simultaneously on the computing platform. After the robot navigates to a BP, it first updates its estimate of the current pose through RGBD  observations, then  the cleaning path  are updated based on the relocation information. An example of a coverage path is shown in Fig.\ref{fig:path}. There are a total of 12 paths, and each side has 6 paths. Finally, the planning algorithm is called to execute the obstacle avoidance task, and the joint space trajectory is obtained. The experimental process is shown in the Fig.\ref{fig:real_exp}, where each picture represents a snapshot of the working situation on one of the BPs shown in Fig.\ref{fig:bps}.

Experiments show all the cleaning paths are successfully planned. However,  paths that are far away from the robot may be partially unreachable. We define the  coverage of  path  as the length of the path performed by the actual robot divided by the length of the total coverage path. In the experiment, the left side achieves 87\% coverage, and the right side is 93\%. If fine-tuning is performed further, which means execute a local base placement search (BPS) nearby the results given by BPO, 100\% coverage can be achieved on both sides.

\section{CONCLUSIONS}
To solve the BPO problem under task uncertainties, this paper uses SLD to model the workspace, which can decouple task constraints and the BPO problem. To obtain optimal coverage, manipulability and task execution time at the same time, this paper formulates the BPO problem as a MOOP. Coverage can be easily calculated using the SLD model. For task execution time, under the full coverage and least time constraints, this paper models it as a TSP problem and the whole problem is solved by an evolutionary algorithm. Finally, based on a real toilet scene, simulation calculations and actual experiments are carried out in this paper. The experiments show that the simulation results can almost be directly transferred to the real scene without further modifications. Furthermore, to ensure the feasibility of the BPs, collision detection is introduced when calculating the FBPs. In order to further improve the optimality of BPO, a local BPS can be  performed. Experiments show that the method proposed in this paper provides a quantifiable and efficient approach for
base placements evaluation and optimization. It avoids on-site debugging through trial and error, which is of great significance for mobile manipulation applications.

After the base placement is determined, the cleaning path is artificially partitioned and assigned to each BP in our experiment. A better practice is to automatically partition the workspace according to the position of BPs, and then generate the cleaning paths corresponding to each partition. Subsequent research will further explore BPO methods combined with automatic partitioning. For the reachability evaluation, this paper uses an offline reachability map, and a learning-based online reachability verification method can also be integrated in the future.



\bibliographystyle{IEEEtran.bst}
\bibliography{root.bib}

\end{document}